\documentclass[10pt,twocolumn]{article}
\usepackage[utf8]{inputenc}
\usepackage{listings}

\usepackage[margin=1in]{geometry}

\AtBeginDocument{%
  \providecommand\BibTeX{{%
    \normalfont B\kern-0.5em{\scshape i\kern-0.25em b}\kern-0.8em\TeX}}}

\usepackage{ifthen}

\makeatletter
\newcommand{\savefont}{\xdef\oldfontsize{\f@size}\xdef\oldblskip{\f@baselineskip}}
\newcommand{\backtoprevfont}{\fontsize{\oldfontsize}{\oldblskip}\selectfont}
\makeatother

\newcommand{\logo}[1]{\savefont{\fontfamily{lmss}\selectfont #1}\backtoprevfont}

\newcommand{\paradiseo}[1][]{%
    \savefont{%
            \ifthenelse{ \equal{#1}{} }%
                {\fontfamily{lmss}\selectfont Paradis\textmd{eo}}%
                {\fontfamily{lmss}\selectfont Paradis\textmd{eo}-\logo{#1}}%
    }\backtoprevfont%
}

\newcommand{\IOH}[1]{\savefont{\fontfamily{lmss}\selectfont IOH{\textmd{#1}}}\backtoprevfont}
\newcommand{\irace} {\savefont{\fontfamily{lmss}\selectfont irace}\backtoprevfont}

\newcommand{\code}[1]   {\lstinline[language=C++]{#1}}

\usepackage{xcolor}

\definecolor{codeback}{rgb}{0.99,0.99,0.99}
\definecolor{codegreen}{rgb}{0.1,0.5,0}
\definecolor{codegray}{rgb}{0.5,0.5,0.5}
\definecolor{codepurple}{rgb}{0.3,0,0.6}
\definecolor{codeblue}{rgb}{0.0,0,0.7}
\definecolor{codered}{rgb}{0.4,0.0,0.0}

\lstdefinestyle{paradiseo}{
    xleftmargin=10pt,
    backgroundcolor=\color{codeback},   
    commentstyle=\rmfamily\color{codegreen},
    keywordstyle=\color{codered},
    morekeywords={size_t},
    numberstyle=\tiny\color{codegray},
    stringstyle=\color{codepurple},
    basicstyle=\footnotesize\ttfamily,
    breakatwhitespace=false,         
    breaklines=true,                 
    captionpos=b,                    
    keepspaces=true,                 
    numbers=left,                    
    numbersep=5pt,                  
    showspaces=false,                
    showstringspaces=false,
    showtabs=false,                  
    tabsize=2
}
\usepackage[colorlinks]{hyperref}
\hypersetup{
    linkcolor=codered,
    anchorcolor=cyan,
    citecolor=codegreen,
    urlcolor=codeblue
}
\usepackage{graphicx}

\title{Paradiseo: From a Modular Framework for Evolutionary Computation to the Automated Design of Metaheuristics\\{\large ---22 Years of Paradiseo---}}

\usepackage{authblk}
\author[1]{Johann Dreo}

\affil[1]{Systems Biology Group,
  Depepartment of Computational Biology,
  USR 3756,
  Institut Pasteur and CNRS,
  Paris, France.
  \texttt{johann@dreo.fr} ---Corresponding author.
  }

\author[2]{Arnaud Liefooghe}
\affil[2]{Univ. Lille, CNRS,
Inria, Centrale Lille,
UMR 9189 CRIStAL,
F-59000 Lille.
\texttt{arnaud.liefooghe@univ-lille.fr}
}

\author[3]{S\'ebastien Verel}
\affil[3]{
  Univ. Littoral C\^ote d'Opale,
  Calais,
  France.
  \texttt{verel@univ-littoral.fr}
}
\author[4]{Marc Schoenauer}
\affil[4]{%
  TAU, Inria, CNRS \& UPSaclay,
  LISN, Saclay,
  France.
\texttt{marc.schoenauer@Inria.fr}
}
\author[5]{Juan J. Merelo}
\affil[5]{%
  University of Granada,
  Granada,   
  Spain.
\texttt{jjmerelo@gmail.com}
}
\author[6]{Alexandre Quemy}
\affil[6]{%
  Poznan University of Technology,
  Poznan,
  Poland.
\texttt{alexandre.quemy@gmail.com}
}
\author[7]{Benjamin Bouvier}
\affil{%
\texttt{public@benj.me}
}
\author[8]{Jan Gmys}
\affil[8]{%
  Inria,
  Lille,
  France.
\texttt{jan.gmys@inria.fr}
}

\date{}

\begin{document}

\maketitle

\begin{abstract}
The success of metaheuristic optimization methods has led to the development of a large variety of algorithm paradigms.
However, {\em no algorithm} clearly dominates all its competitors on {\em all problems}. Instead, 
the underlying {\em variety of landscapes} of optimization problems calls for
a variety of algorithms to solve them efficiently.
It is thus of prior importance to have access to mature and flexible software frameworks
which allow for an efficient exploration of the {\em algorithm design space}.
Such frameworks should be flexible enough to accommodate any kind of metaheuristics,
and open enough to connect with higher-level optimization, monitoring and evaluation softwares.
This article summarizes the features of the \paradiseo{} framework,
a comprehensive C++ free software which targets the development of modular metaheuristics.
\paradiseo{} provides a highly {\em modular} architecture, a large set of {\em components},
{\em speed} of execution and {\em automated algorithm design} features,
which are key to modern approaches to metaheuristics development.
\end{abstract}



\section{Introduction}

In the research domain of metaheuristics for black-box optimization,
a very large variety of algorithms has been developed since the first {\em Evolution Strategies} appeared in 1965~\cite{RECHENBERG1965}.
Starting from nature-inspired computing methods and following recent mathematical approaches, numerous applications have shown the efficiency of those randomized search heuristics.
However, following Wagner et al.~\cite{Wagner2014}, we observe that the metaheuristic research domain lacks mature software,
while it is crippled with short-lived research prototypes on over-specific features sets.
We believe this state hinders the adoption of those technologies in the industrial world
and is an obstacle to breakthrough innovations.
Therefore, the development of a full-featured and mature metaheuristic optimization framework is of prior importance,
for both the scientific and the applied communities.
In this article, we 
summarize our 
efforts towards this goal,
in the guise of the \paradiseo{} project.

The \paradiseo{} framework is a 22 years old effort which aims at developing a flexible architecture
for the generic design of metaheuristics for hard optimization problems.
It is implemented in C++, a very mature object-oriented programming language,
which is probably one of the fastest, if not \emph{the} fastest, object-oriented programming platforms on the market \cite{Hundt2011,nesmachnow2015empirical,merelo2017ranking}. It is also highly portable and benefits from very extensive tooling as well as an active community.
\paradiseo{} is released as a free and open-source software, under the LGPL-v2 and CeCILL licenses (depending on the module).
Its development is open and the source code is freely available on the
Inria\footnote{\url{https://gitlab.inria.fr/paradiseo/paradiseo}}
and Github\footnote{\url{https://github.com/jdreo/paradiseo}}
code repositories.

\subsection{History}

The ``Evolving Objects'' (\logo{EOlib}, then simply \logo{EO}) framework was started in 1999 by the Geneura team at the University of Granada, headed by Juan Julián Merelo.
The development team was then  reinforced by Maarten Keijzer, who designed the current modular architecture, and Marc Schoenauer \cite{Keijzer2001}.
Later came Jeroen Eggermont, who, among other things, did a lot of work on genetic programming, Olivier König,
who did a lot of useful additions and cleaning of the code, and Jochen Küpper.

The Inria {\it Dolphin} team, headed by El-Ghazali Talbi, did a lot of contributions starting from around 2003, on their own module collection called \paradiseo{}.
Thomas Legrand worked on particle swarm optimization, the regretted Sébastien Cahon and Nouredine Melab
worked on parallelization modules~\cite{Cahon2003,Cahon2003a,Cahon2004,Cahon2004a}.
Arnaud Liefooghe and Jérémie Humeau worked a lot on the multi-objective module~\cite{Liefooghe2011} 
and on the local search one along with Sébastien Verel~\cite{Humeau2013}.
In the same team, C. FC.\footnote{Redacted by author's demand.} 
and Jean-Charles Boisson made significant contributions.

The (then) \logo{EO} project was taken over by Johann Dreo, who worked with the help of Caner Candan on adding the \logo{EDO} module.
Johann and Benjamin Bouvier have also designed a MPI parallelization module, while Alexandre Quemy also worked on parallelization code.

In 2012, the two projects (\logo{EO} and \paradiseo{}) were merged into a single one by Johann Dreo, Sébastien Verel and Arnaud Liefooghe,
who have been acting as maintainers ever since.

In 2020, automated algorithm selection design and binding toward the \IOH{profiler} validation tool were added by Johann Dreo.


Along the life of the project, several spin-off software have been developed,
among which a port of the EO module in Java~\cite{arenas2002jeo},
another one in ActiveX\footnote{\url{http://geneura.ugr.es/~jmerelo/DegaX/}};
GUIDE, a graphical user interface for assembling algorithms~\cite{Collet2004}\footnote{Which also supported ECJ.},
and EASEA, a high-level declarative language for evolutionary algorithm specification~\cite{Collet2000},
which later became independent~\cite{Maitre2012} of the specific library.

\subsection{Related Frameworks}

The 1998's version of the hitch-hiker’s guide to evolutionary computation
(frequently asked question in the \texttt{comp.ai.genetic} Usenet
newsgroup\footnote{Discussion forum which was popular before the World Wide Web and social networks. \url{http://coast.cs.purdue.edu/pub/doc/EC/FAQ/www/Q20.htm}})
already lists 57 software packages
related to the implementation of evolutionary algorithms (among which \logo{EOlib}, the ancestor of \paradiseo{}).

Most of those software are now unmaintained or impossible to find.
There has been, however, a constant flow of new frameworks, library or solvers every year, for decades.
We were able to find at least 47 of them readily available on the
web\footnote{\paradiseo{}, jMetal, ECF, OpenBeagle, Jenetics, ECJ, DEAP, CIlib, GP.NET, DGPF, JGAP, Watchmaker, GenPro, GAlib, HeuristicLab, PyBrain, JCLEC, GPE, JGAlib, pycma, PyEvolve, GPLAB, Clojush, µGP, pySTEP, Pyvolution, PISA, EvoJ, Galapagos, branecloud, JAGA, PMDGP, GPC++, PonyGE, Platypus, DCTG-GP, Desdeo, PonyGE2, EvoGrad, HyperSpark, Nevergrad, Pagmo2, LEAP, Operon, EMILI, pso-de-framework, MOACO.}.
Among those projects, only 8 met all the following criteria:\begin{enumerate}
    \item open-source framework aiming at {\em designing} algorithms\footnote{{\em Libraries} of solvers, like Pagmo2 or Nevergrad, do not match this criterion.},
    \item being active since 2015,
    \item having more than 15 contributors.
\end{enumerate}
The features of those main frameworks are compared in Table~\ref{table:frameworks}, where the number of lines of code was computed with the \texttt{cloc} tool\footnote{version 1.82 of \url{https://github.com/AlDanial/cloc}}.
Note that for \logo{HeuristicLab}, the code for the GUI modules was excluded from the count.
The GPGPU module of \paradiseo{}~\cite{Melab2013} is not counted either, as it is not maintained anymore.
The number of contributors has been retrieved from the code repository's commit histories,
which underestimates the number of people involved in the case of \paradiseo{};
the extracted number is however kept, for fairness in comparison with the other frameworks, that might face a similar bias.

\begin{table}[!t]
    \caption{Main software frameworks for evolutionary computation and metaheuristics.
    Fastest languages are figured in green and slowest in red,
    copyleft licenses are in red. ``kloc'' stands for ``thousands of lines of code''.}
    \label{table:frameworks}
    \centering
    \includegraphics[width=1.0\columnwidth]{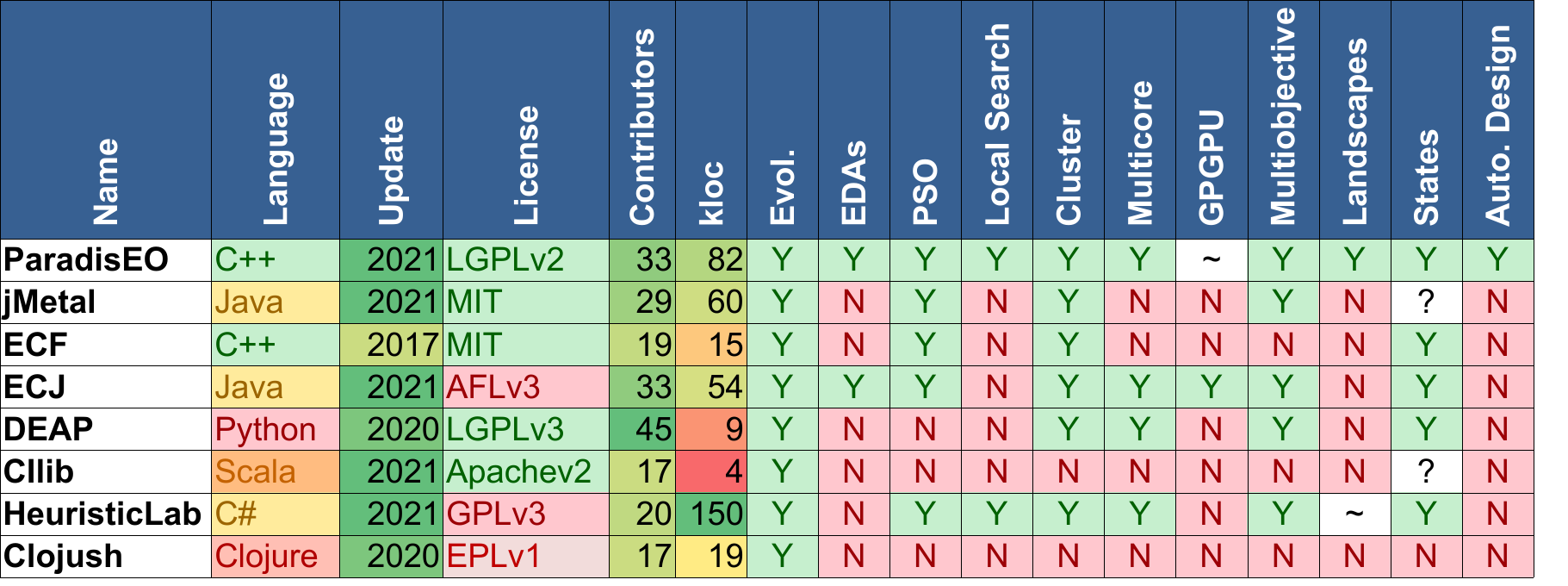}
\end{table}

Among the software close in features to \paradiseo{}, \logo{ECF} has not been updated in 4 years.
\logo{ECJ}, \logo{jMetal} are close competitors, albeit programmed in Java, which is expected to run near 2.6 times slower than programs in C++\footnote{Following the "n-body" setup of the "Computer Language Benchmarks Game", which is the closest problem to our setting: \url{https://benchmarksgame-team.pages.debian.net/benchmarksgame/performance/nbody.html}},
a key drawback for automated algorithm design (see Section~\ref{sec:auto-design}).
\logo{HeuristicLab} suffers from the same drawback,
but provides a graphical user interface for the run and analysis of solvers.
\paradiseo{} does not provide such a GUI, but relies on dedicated third-party tools for this kind of functionality
(see~\cite{Collet2004} and Section~\ref{sec:auto-design}).
The other frameworks do not provide the same level of features and use languages that are generally slower than C++~\cite{merelo2017ranking}.

\section{Architecture}

From its inception \cite{EO:FEA2000}, \paradiseo{} opted for an original architecture design, exemplified by its name, ``Evolving {\em object}'', as opposed to a procedural or functional view of the algorithm. In this section, we expose first the main concepts used in its architecture, to focus next on the design patterns that have been used in it, giving it room for evolution and improvement along the years.

\subsection{Main Concepts}

The core of \paradiseo{} is formed by the \logo{EO} module, which has been designed for general evolutionary algorithms.
Most of its core concepts are used across the other modules and are named after its vocabulary:
\begin{description}
    \item[Encoding:]   The data structure modelling a solution to the optimization problem (which type is generally denoted \code{EOT}).
    \item[Evaluation:] The process of associating a value to a solution, thanks to an objective function.
    \item[Fitness:]    The value of a solution as seen by the objective function.
    \item[Operator:]   A function which reads and/or alters a (set of) solutions.
    \item[Population:] A set of solutions.
\end{description}

\subsection{Main Design Patterns}

\begin{figure}[htbp]
    \centering
    \includegraphics[width=0.6\columnwidth]{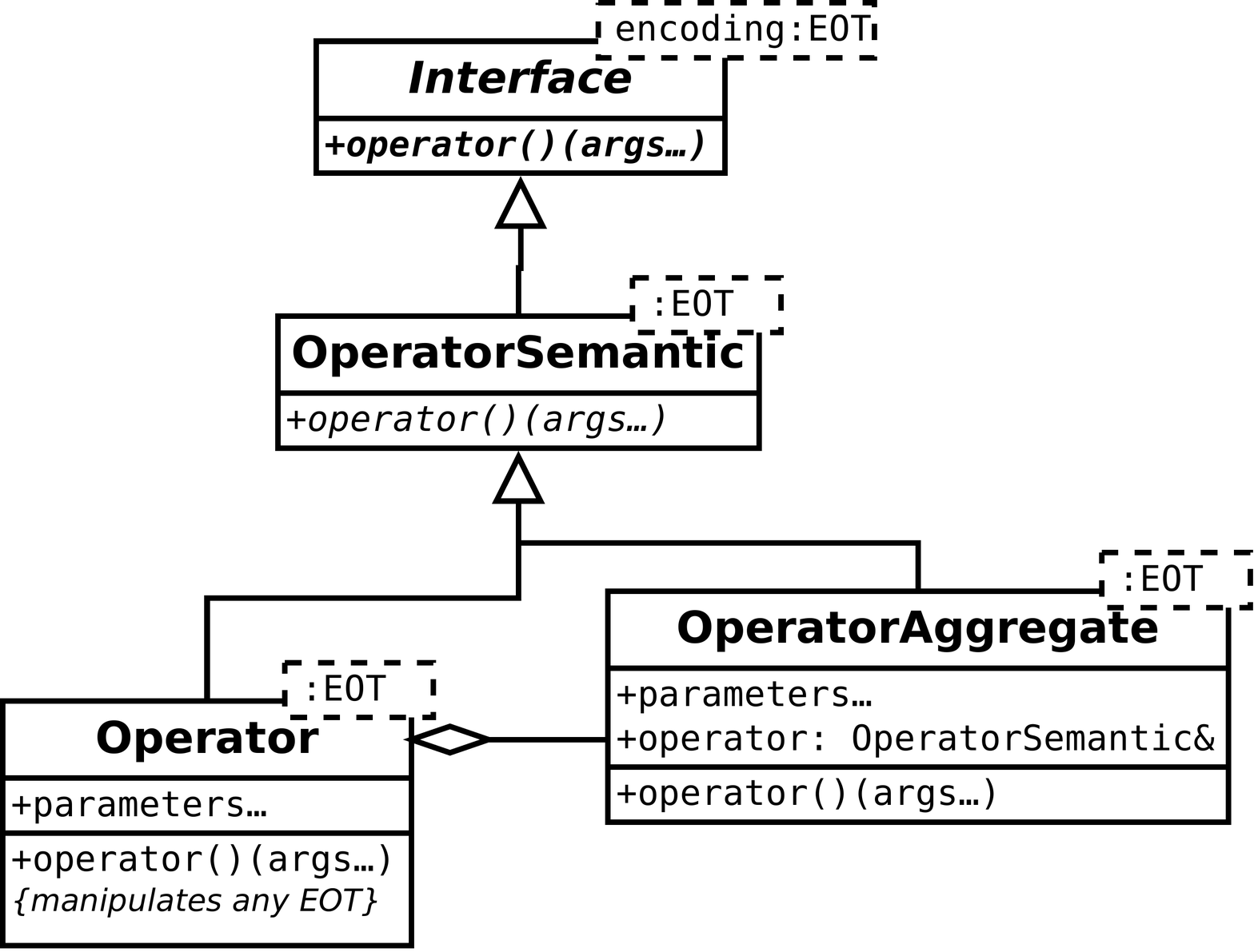}
    \caption{UML diagram of a high-level view of the main design pattern used in the \paradiseo{} framework.}
    \label{fig:main_design_pattern}
\end{figure}

\paradiseo{} is a {\em framework}, providing a large set of components that the user can assemble to implement a solver.
To facilitate and enforce the design and use of components, \paradiseo{} is based on four main design patterns: Functor, Strategy, Generic Type and Factory.
Figure~\ref{fig:main_design_pattern} shows a high-level view of the global design pattern.

\begin{description}
    \item[Functor:] In \paradiseo{}, most of the operators are {\em functions} (exposing the 
    \code{operator()} interface) holding a {\em state} between calls~\cite{Vandevoorde2003,Keijzer2001}.
    Member variables of the functors are either parameters or references to other functors, involved in the computation.
    
    \item[Strategy:] Operators can be composed to form another one. For instance, an \code{eoAlgo} is essentially an operator holding a loop which calls other operators. These operators must then honor an interface, which provides the {\em semantic} of the underlying operation~\cite{gamma1995design}. For instance, an \code{eoSelectOne} exposes the interface to pick a solution within a {\em population}: \code{const EOT& operator()(const eoPop<EOT>&)}.
    
    \item[Generic type:] Almost all the operators in \paradiseo{} are defined over a \code{EOT} template holding the {\em encoding} of a solution to the optimization problem. This allows for two crucial features: (i) the user can provide her own data structure, without major redesign, and (ii) any operator deep in the call tree may have access to any specific interface of the encoding~\cite{Vandevoorde2003}. This was one of the earliest decisions taken, and was already presented in \cite{Keijzer2001}.
    
    \item[Factory:] As many operators are abstracted through their interfaces,
    \paradiseo{} provides ways to manage them as collections or high-level aggregates,
    so that the user does not have to manage the details.
    For instance, \paradiseo{} provides classical stopping criterion collections
    or on-the-fly instantiation (see also Section~\ref{sec:auto-design}).
\end{description}
With this approach, \paradiseo{} is enforcing the use of {\em composition} of objects, limiting the use of {\em inheritance} to interfaces (generally of abstract classes).
One of the main goals of the framework is to be able to easily {\em compose} new algorithms,
by (i) reusing existing common features (logging, parallelization, state serialization, etc.) and
(ii) assembling existing algorithmic operators, for instance by hybridizing algorithms.

\begin{figure*}
    \centering
    \includegraphics[width=0.9\textwidth]{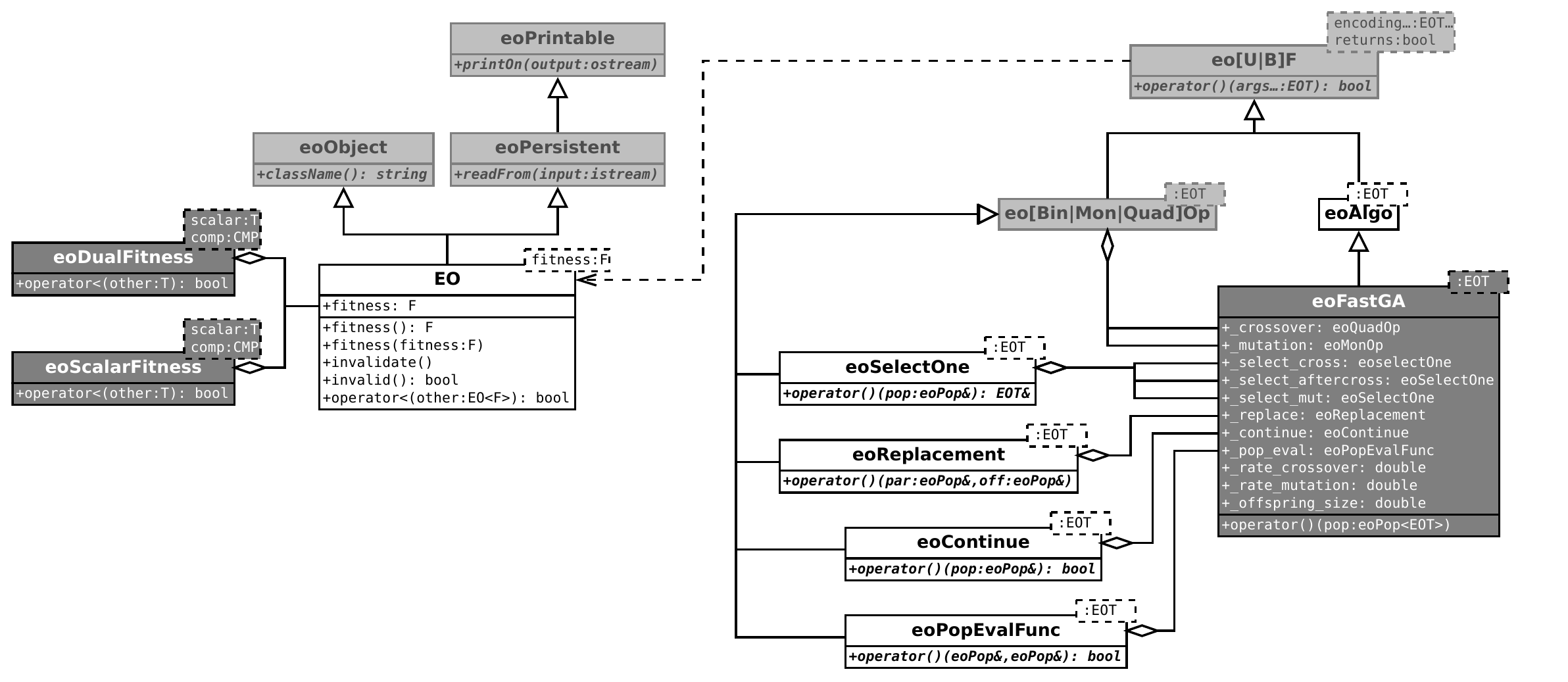}
    \caption{General overview of the main classes involved in assembling one of the \paradiseo[EO] algorithms.
    Concrete class implementations manipulated by the user are shown in dark gray.
    Interfaces which define generic behavior are figured in white.
    Low-level convenience classes provided by the framework are figured in light gray.}
    \label{fig:overview_algo}
\end{figure*}

Figure~\ref{fig:overview_algo} shows an example of the core concepts which a user may see: the \code{EO} class, from which a solution to the optimization problem will inherit the interface necessary to be used across the operators, and a set of operator interfaces and implementations, which define an algorithm.

\section{Modules}

\paradiseo{} targets modular algorithms and is thus organized in several modules:
Evolving Objects (\logo{EO}) for population-based algorithms such as evolutionary algorithms and particle swarm optimization,
Moving Objects (\logo{MO}) for local search algorithms and landscape analysis,
Estimation of Distribution Objects (\logo{EDO}) for estimation of distribution algorithms, and
Multi-Objective Evolving Objects (\logo{MOEO}) for multi-objective optimization.
Each module brings its own specific concepts and features, on top of the common core features provided in the \logo{EO} module.
The following sections summarize this organization.

\subsection{Evolutionary and Particle Swarm Algorithms --- \logo{EO}}

The \logo{EO} module defines a number of common operators which are used across all the framework:
\begin{description}
    \item[Evaluation of populations:] Operators which call the objective function for a set of solution. With several parallelization options~\cite{Cahon2003,Cahon2003a,Cahon2004,Cahon2004a}.
    \item[Initialization:] Operators which generate solutions, out of the optimization loop (generally at random).
    \item[Continue:]    A stopping criterion which returns \code{false} if the iteration loop of an algorithm is to be stopped.
    \item[Checkpoints:] A generic operator which is called at each iteration (for instance to collect statistics).
    \item[Updater:]     A generic operator which can update a parameter.
    \item[State:]       A serialization of a state of an algorithm.
    \item[Wrappers:]     Operators which transform a (set of) operators as another one.
    \item[Generic encodings:] Some solution representations which are often used when solving optimization problems, like numerical vectors, binary vectors, trees, permutations.
    \item[Parameters:] An abstraction of parameters, which can be used for command line interfaces, state management and dynamic algorithms.
\end{description}
Those features are generally used by the following modules.


The \logo{EO} module holds the following necessary classes to implement evolutionary algorithms, as illustrated in Figure~\ref{fig:EDO_loop} (lower part):
\begin{description}
    \item[Selection:] Operators which pick (a set of) solutions within a population. Two levels are available: operators which pick a single solution (\code{eoSelectOne} interface), and operators which select 
    more than one solution (\code{eoSelect}).
    \item[Variation:]  Operators which generate new solutions by altering existing ones. They are generally called ``mutations'', when they alter a single solution, and ``crossover'' when they alter two or more solutions at once. Both types return a boolean which is \code{true} if an alteration has actually been done.
    \item[Replacement:] Operators which merge two populations, typically ``parents'' with ``offsprings'' (produced by the alteration of parents with variation operators).
    \item[Algorithms:] High-level operators which manipulate a population generated at initialization, and iteratively apply a set of operators until a stopping condition is satisfied.
\end{description}
More information on the design of evolutionary algorithms within \paradiseo[EO] can be found in~\cite{Keijzer2001}.

\logo{EO} also defines classes which target particle swarm optimization algorithms:
\begin{description}
    \item[Particle:] An interface on top of the \code{EO} class, which defines a freely moving particle.
    \item[Velocity:] Operators which control the speed at which a particle is moving.
    \item[Flight:] Operators which control the next position at which the particle will be.
\end{description}

\subsection{Local Search and Landscape Analysis --- \logo{MO}}

The \logo{MO} module adds an interface which can manage {\em single solutions} instead of {\em populations}, mainly providing a fine-grained level of abstraction, following the same components as the \logo{EO} module.
It also adds the important concept of {\em incremental evaluation},
to allow the design of objective functions which compute the value of a solution based on
the application of a {\em move} to an already evaluated solution.
The objective function can thus take into account only the sub-parts of the solutions that have been altered,
effectively improving the computation time.
Those operators are tightly coupled with {\em neighborhoods}, which are variation operators applied to single solutions.

The \logo{MO} module additionally provides components to sample the search space and estimate statistics for characterizing the \emph{fitness landscape} of the problem in terms of features, such as the density of states, the fitness distance correlation, the autocorrelation function, the length of adaptive walks, the landscape neutrality, or the fitness cloud~\cite{hoos2004,tuto_landscape}.
More information about local search and landscape analysis in \paradiseo[MO] can be found in~\cite{Humeau2013}.

\subsection{Multi-Objective Optimization --- \logo{MOEO}}

The \logo{MOEO} module adds the necessary features to handle multi-objective optimization~\cite{coello2007,zitzler2004}:
\begin{description}
    \item[Fitness assignment:] Large set of operators which convert raw objective values into ranks or fitness values used for selection and replacement.
    They include state-of-the-art scalarizing-, 
    dominance- and indicator-based approaches.
    \item[Diversity preservation:] Operators which maintain diversity in the population, seeking for well-spread and uniformly-distributed solutions in the objective space.
    \item[Selection:] Operators which combine the ones above in order to guide the population towards Pareto-optimal solutions.%
    \item[Archive:] Secondary population which maintains non-do\-mi\-na\-ted solutions.
    \item[Performance metrics:] Quality indicators 
    computed over populations or archives to measure solution quality in multi-objective optimization.
\end{description}
More information about the \paradiseo[MOEO] module, and how to design multi-objective local search and evolutionary algorithms in \paradiseo{} are detailed 
in~\cite{Liefooghe2011}. 

\subsection{Estimation of Distribution --- \logo{EDO}}

\begin{figure}
    \centering
    \includegraphics[width=0.8\columnwidth]{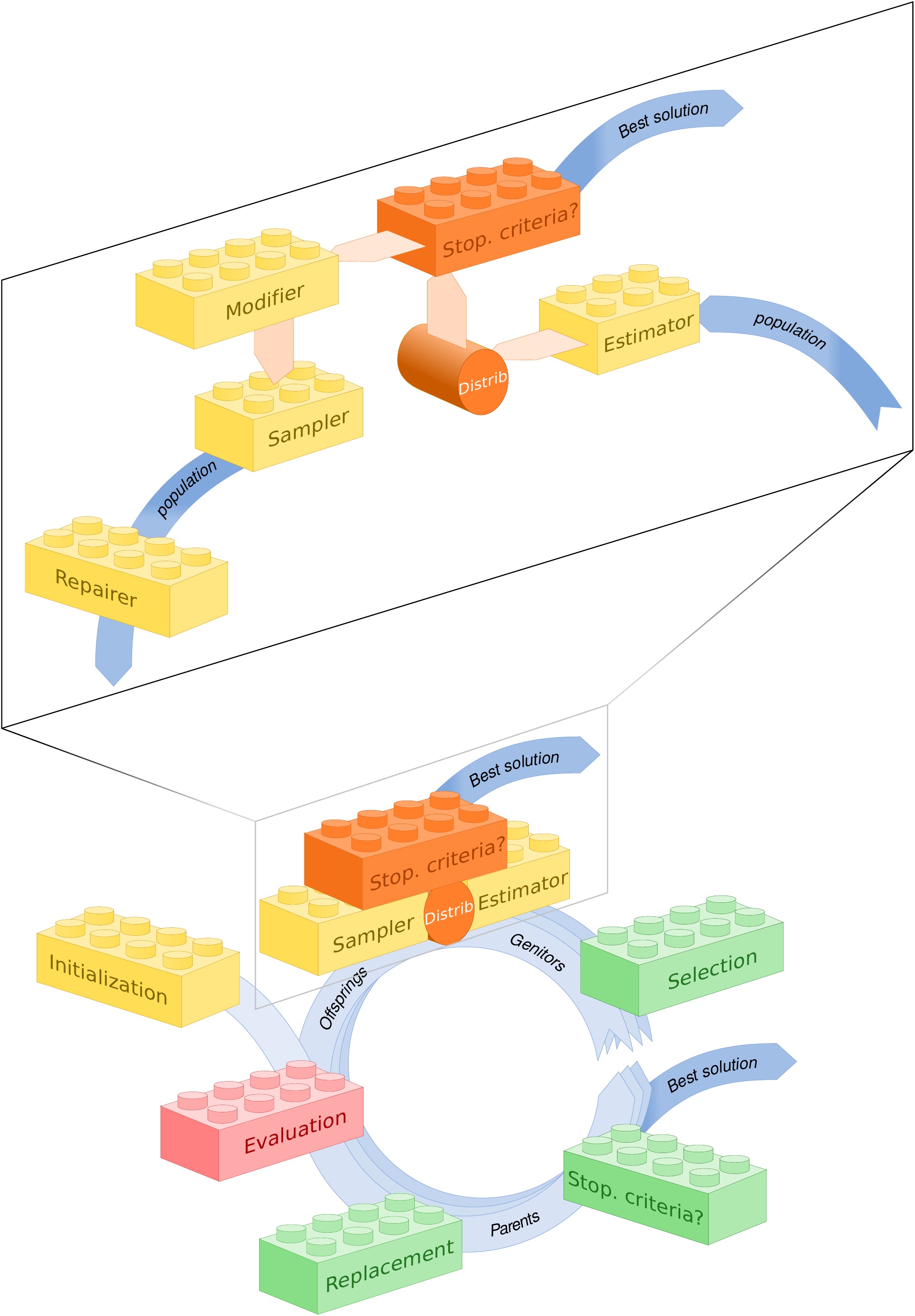}
    \caption{Modular estimation of distribution algorithm as seen from \paradiseo[EDO]. The lower part of the diagram is the modular evolutionary algorithm loop.
    \logo{EDO} adds a set of operators to replace ``implicit'' variation operator by ``explicit'' ones. The operators managing the probability distribution are shown in orange.}
    \label{fig:EDO_loop}
\end{figure}

The \logo{EDO} module encompasses the features to manage population-based algorithms
which have an explicit state from which the population is derived at each iteration.
\begin{description}
    \item[Distribution:] A template, wrapping the encoding \code{EOT} and holding the data structure representing a probability distribution.
    \item[Estimator:] Operators which compute distribution parameters from a given population.
    \item[Sampler:]   Operators which compute a population from a given distribution.
\end{description}
Several other operators allow to manipulate and combine those objects and to plug them
within \logo{EO} evolutionary algorithm's variation operator, as shown on Figure~\ref{fig:EDO_loop}.

\section{Key Features\label{sec:key-features}}

\subsection{Modular Algorithms\label{sec:modular}}

\begin{figure*}
    \centering
    \includegraphics[width=0.85\textwidth]{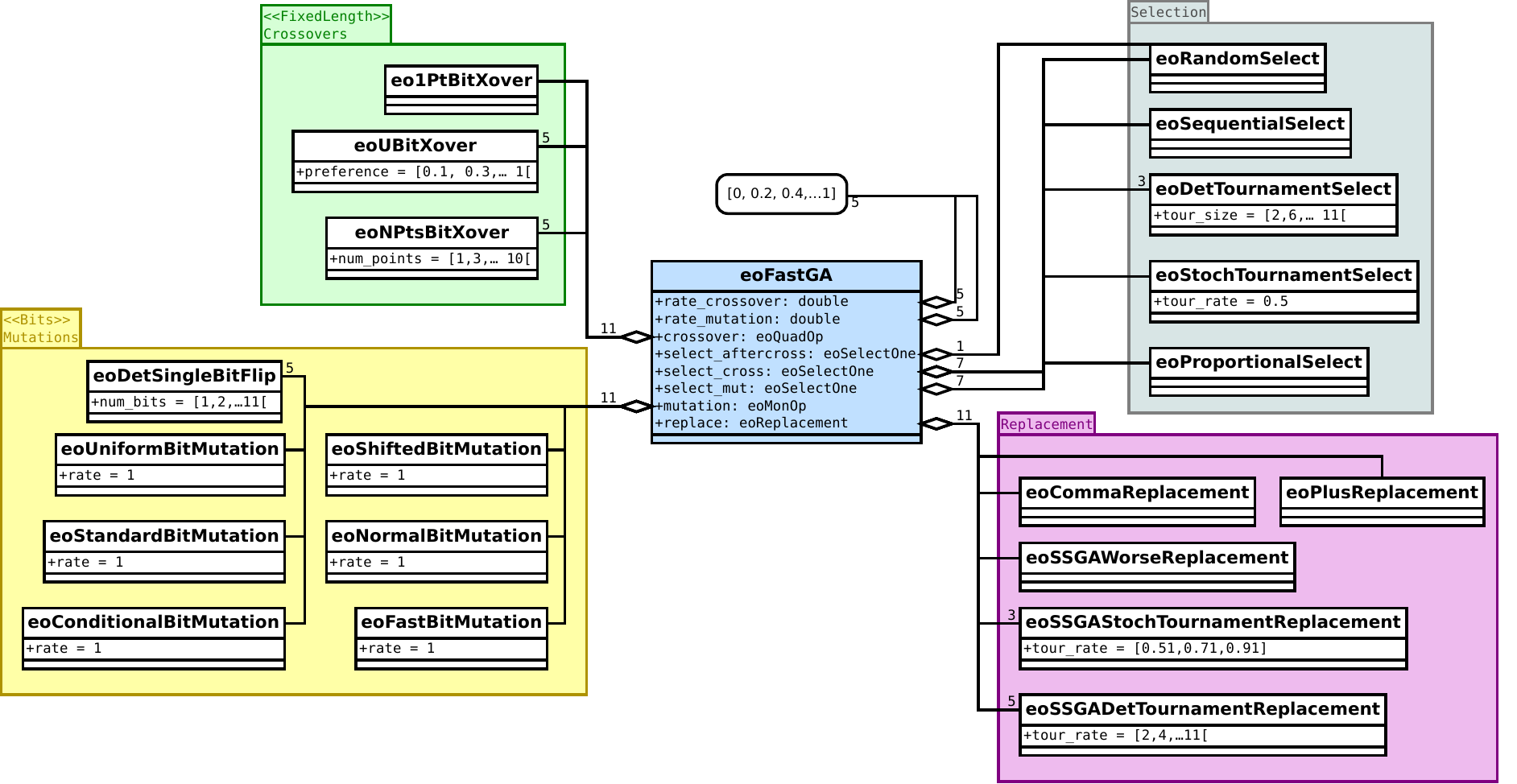}
    \caption{Example of relationships between an algorithm template (\code{eoFastGA}) and its related operators. Each package box groups alternative operators which may be used for the corresponding step of the algorithm.}
    \label{fig:FastGA}
\end{figure*}

The main feature of \paradiseo{} is to provide a large set of modular algorithms,
which are assembled from a large variety of operators.
This is motivated by the fact that there exists a large diversity of optimization problems,
which would be more efficiently solved by specific algorithms rather than generic ones.

It is thus of prior importance to be able to easily explore the design space of algorithms,
in order to find the best one for a given problem.
Having a large set of reusable components is key to allow the practitioner to quickly try
new algorithm variants, which may not have been tested yet. 
New ideas can also be experimented with minimum effort, by allowing the user to focus on a single (new) component.
Figure~\ref{fig:FastGA} shows a simple example of a modular genetic algorithm (inspired from~\cite{doerr2013lessons}),
which allows for the instantiation of $1\,630\,475$ different algorithm {\em instances}.
In that case, an algorithm instance is a combination of parameterized operators,
with varying functions and/or parameters\footnote{Generally numerical or integer parameters, sometimes boolean or categorical ones.}.
Of course, considering the whole footprint of \paradiseo{} would allow for far larger design space.

It is also worth noting that the hybridization of two algorithms in \paradiseo{}
is as simple as encapsulating operators with a similar interface.
For instance, it is straightforward to use a local search algorithm implemented with \logo{MO}
as a variation operator of an evolutionary algorithm implemented with \logo{EO}, then ending up with a so-called memetic algorithm.

Moreover, this modular architecture facilitates a fair comparison of algorithms in practical use cases,
where {\em wall-clock} performance is of prior importance;
e.g., for applications involving interactions with a human.
In such a case, having a common code base helps conducting more unbiased studies.

\subsection{Fast Computations\label{sec:fast}}
\paradiseo{} is one of the few optimization frameworks written in C++, a compiled programming language known for its runtime speed.
Moreover, its design is thought to directly plug components at compile time rather than relying exclusively on dynamically-run conditional expressions.

A typical rationale in black-box optimization is to state that the efficiency of the algorithm computations
is not a concern, because in real cases the objective function dominates the runtime.
While this is true in essence, this argument forgets that, during the design phase of algorithms,
practitioners most often do not use complex objective functions, but synthetic ones, which are very fast to compute.
In that case, fast computation means fast design iterations.

For example, a CMA-ES algorithm implemented with \paradiseo{} is 10 times faster
than its heavily optimized counterpart implemented with Python/Numpy,
when solving a standard synthetic benchmark.
Those measures are obtained using the reference implementation of CMA-ES available in the \logo{pycma} package\footnote{Version 3.0.0 of \url{https://github.com/CMA-ES/pycma}},
solving the Black-Box Optimization Benchmark
(BBOB~\cite{Hansen2010}) of the COmparing Continuous Optimizers (\logo{COCO}\footnote{Version 2.3.2 of \url{https://github.com/numbbo/coco}}) platform~\cite{Hansen2021}.
The \paradiseo{} implementation\footnote{Version
\href{https://github.com/nojhan/paradiseo/commit/640fa31fb510d620cc0aef069dce2c615aee2a80}{640fa31} of \url{https://github.com/nojhan/paradiseo}}
used the independent implementation of BBOB available on the \IOH{experimenter}\footnote{Version \href{https://github.com/nojhan/IOHexperimenter/commit/2395af46ca23273b800978c41dde0196039fe16e}{2395af4} of \url{https://github.com/nojhan/IOHexperimenter}}
platform~\cite{Doerr2018}. Both benchmarks are implemented in C/C++.
Running algorithms on the whole benchmark, on a single Intel Core i5-7300HQ at 2.50GHz with a Crucial P1 solid-state disk, takes approximately 10 minutes with \logo{pycma}/\logo{COCO}, and only 1 minute with the \paradiseo{}/\IOH{} implementation.

In addition, the Symmetric Multiprocessing module (\logo{SMP}) allows to wrap any operator called within a loop transparently to fully make use of CPU cores. The master-worker model has been shown to scale (near) linearly with the number of cores, while having a low communication overhead. \logo{SMP} also provides a parallel island model~ \cite{whitley1999island} that speeds up algorithm convergence while maintaining diversity.

At last, we argue that fast computations of the algorithm and objective function are necessary features
to facilitate automated algorithm design, where an algorithm is itself in charge of finding the most
appropriate variant of an algorithm~\cite{Kerschke2019},
{\em learn} what is the best algorithm for a given benchmark~\cite{Hutter2006},
or even a given problem {\em fitness landscape}~\cite{LeytonBrown2002,belkhir2017,tuto_landscape}.
In that case, running an assembled algorithm on a benchmark is the objective function
of the design problem, and its computation time determines the scale at which the experiment can be conducted. 
This feature is discussed in more detail below.

\subsection{Automated Algorithm Design\label{sec:auto-design}}

\begin{figure*}
    \centering
    \includegraphics[width=0.7\textwidth]{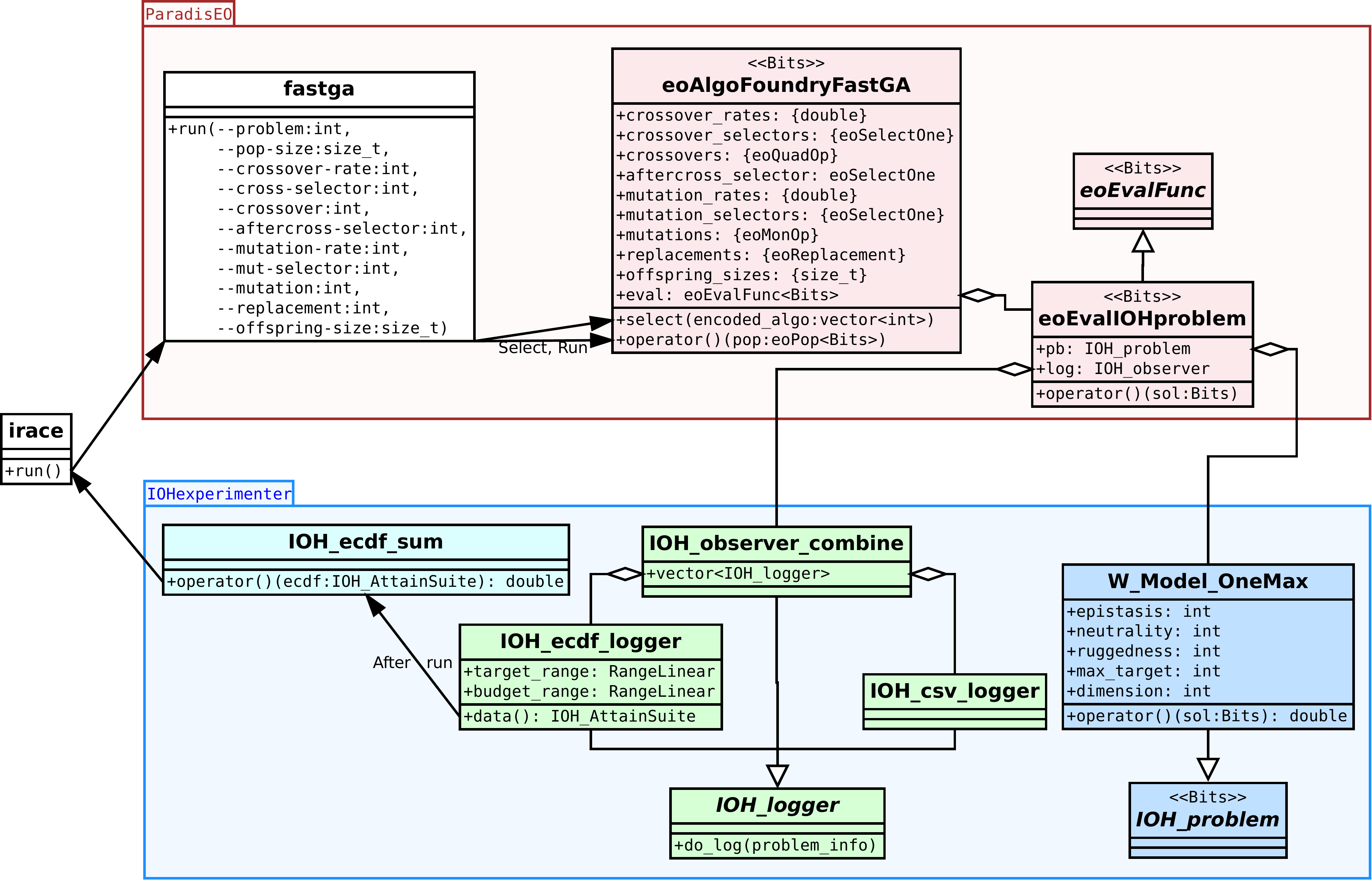}
    \caption{Flow of information involved in automated algorithm design,
    starting with \irace{} calling an executable interface (white box),
    to instantiate and run a \paradiseo{} algorithm (red boxes),
    which will call an \IOH{experimenter} problem (blue boxes)
    while being observed by a logger (green boxes).
    The final performance is computed (cyan box) and returned to \irace{}.}
    \label{fig:IOH_UML}
\end{figure*}

Automated algorithm design features are recent additions to \paradiseo{}.
They target the ability to assemble algorithms at runtime without loss of performance,
and easy bindings with benchmarking and algorithm selection tools.
Figure~\ref{fig:IOH_UML} shows the global setting, which is detailed in this section.

\begin{figure}[htbp]
    \centering
    \includegraphics[width=0.9\columnwidth]{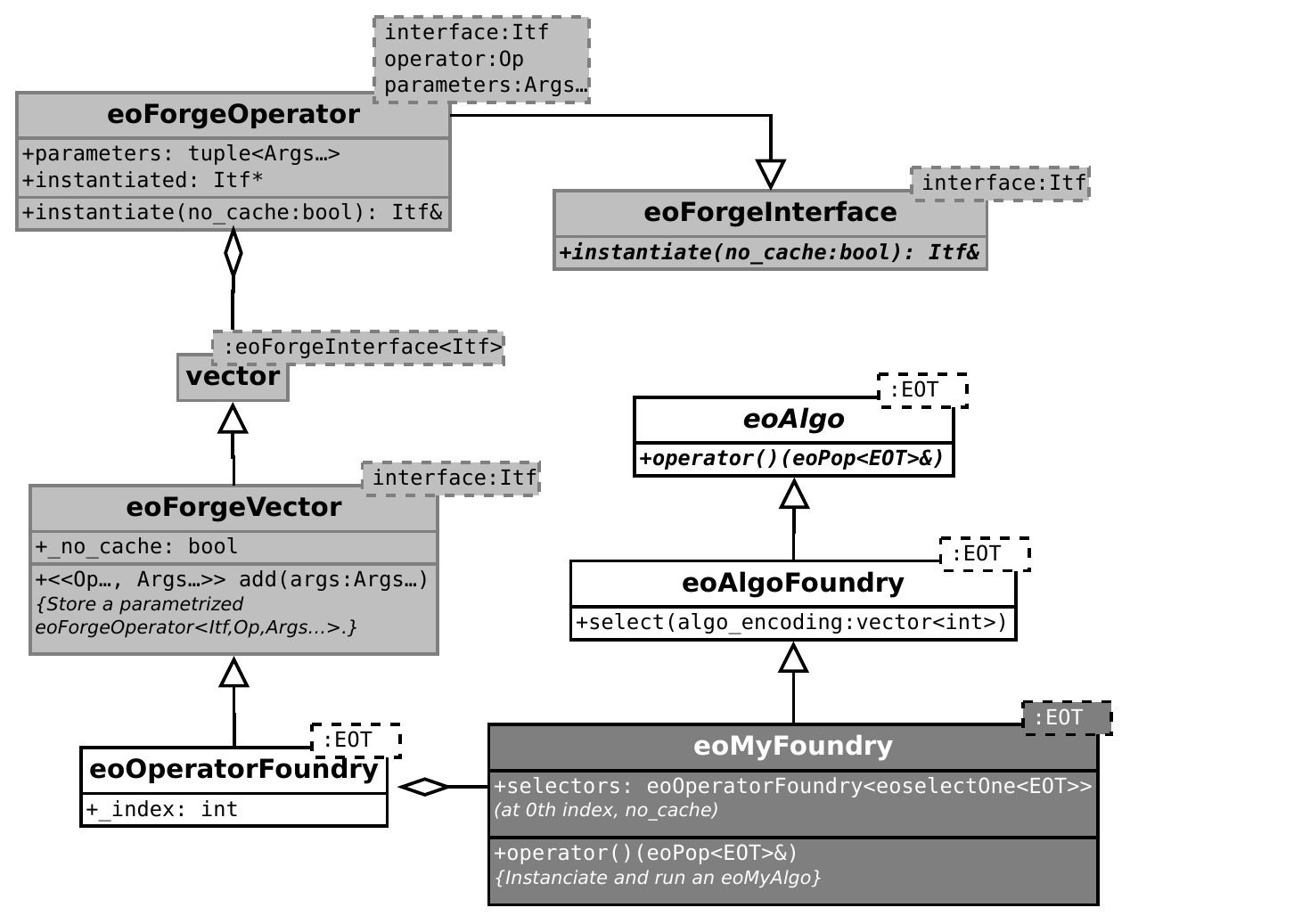}
    \caption{Classes involved in a meta-algorithm instantiation. The \code{eoMyFoundry} class is to be designed by the practitioner, who has to know the interfaces presented with a white background. Classes with a grey background are the underlying framework machinery.}
    \label{fig:foundries}
\end{figure}

\paragraph{On-the-fly Operator Instantiation with Foundries}
Foundries are ``Factory'' classes which allow to instantiate a parameterized operator,
chosen among a set of operators having the same interface.
The user can indicate which classes and parameters should be managed
and a foundry is responsible for instantiating when it is called
with the index of the operator to be instantiated.
This allows for simple numerical interfaces with algorithm selection solvers (i.e. generic hyper-parameters tuning).
An algorithm foundry is thus a generic meta-algorithm, which can instantiate and call an actual algorithm class.
It follows the same interface as an algorithm, but models the operators of this algorithm as {\em operator foundries} rather than references to operator instances.
Those operator foundries are responsible for instantiating the operator when asked to do so.
Figure~\ref{fig:foundries} shows the classes involved.

An end-user willing to find the best algorithm variant for her needs would need to select a subset of parameterized operators of interest and add them to the foundry.
Then, it is sufficient to indicate (potentially at runtime) which algorithm should be instantiated and run.
Let us illustrate this with an example in Listing~\ref{lst:foundry}.

\begin{lstlisting}[language=C++,label=lst:foundry,caption={Excerpt of the use of an algorithm foundry.}]
// Considering the FastGA modular algorithm,
// solving a problem with fixed initialization.
auto& foundry = store.pack< eoAlgoFoundryFastGA<Bits> >(
  init, problem, max_eval_nb, /*max_restarts=*/1);
// Consider different crossover operators.
for(double i=0.1; i<1.0; i+=0.2) {
  foundry.crossovers.add< eoUBitXover<Bits> >(i);
  foundry.crossovers.add< eoNPtsBitXover<Bits> >(i*10);
}
// And different variation rates.
for(double i=0.0; i<1.0; i+=0.2) {
    foundry.crossover_rates.add<double>(i);
    foundry.mutation_rates.add<double>(i);
}
// etc.
// Decide which operators to use.
Ints encoded_algo(foundry.size());
encoded_algo[foundry.crossovers     .index()] = 2;
encoded_algo[foundry.crossover_rates.index()] = 1;
encoded_algo[foundry.mutation_rates .index()] = 3;
// etc.
// Instantiate the operators, or use cached objects.
foundry.select(encoded_algo);
// Run the selected algorithm.
eoPop<Bits> pop; // [...]
foundry(pop);
\end{lstlisting}

\paragraph{Binding with \IOH{} for Fast Benchmarking}
One of the key features when doing automated algorithm design is the ability
to run the assembled algorithm against a whole benchmark,
and to measure performance on this experiment.
\paradiseo{} is not intended to host benchmarks, apart for a few examples,
but provides an interface to the \IOH{experimenter} benchmarking platforms.
\IOH{experimenter} provides a set of benchmarks, along with a standardized logging system,
which can log both calls to the objective function and parameters status.
\paradiseo{} provides several entry points to \IOH{experimenter} problems,
in the form of sub-classes of the \code{eoEvalFunc} interface,
which can be plugged into any \paradiseo{} algorithm.
\IOH{experimenter} also provides a way to extract performance measures from
the runs' logging outputs, with statistics computed on aggregated in-memory
discrete empirical cumulative density functions.
Those distributions are defined on both the computation time
and the quality of solutions axes and can thus produce many
different performance metrics.
This allows for a fast logging and performance assessment system,
which is ideal for automated algorithm design.
Listing~\ref{lst:IOH} shows an example of use of the
\paradiseo{}/\IOH{} binding when solving a 
particular problem.
\\

\begin{lstlisting}[language=C++,label=lst:IOH,caption={Excerpt of the use of the \IOH{} binding.}]
// In-memory logger.
IOHprofiler_RangeLinear<size_t>
  target_range(0, max_target, buckets),
  budget_range(0, max_evals , buckets);
IOHprofiler_ecdf_logger<int,int,int> ecdf_logger(
  target_range, budget_range);
// Benchmark problem.
W_Model_OneMax w_model_om;
ecdf_logger.track_problem(w_model_om);
// The actual Paradiseo/IOH interface:
eoEvalIOHproblem<Bits> pb(w_model_om, ecdf_logger);
// [`pb` is plugged into an algorithm and ran...]
// The performance of the run is recovered:
IOHprofiler_ecdf_sum ecdf_sum;
long perf = ecdf_sum(ecdf_logger.data());
\end{lstlisting}
It is worth noting that the same approach 
can be used with the \IOH's file logger,
which allows for a fine-grained analysis of the algorithm behavior within the
\IOH{analyzer} graphical user interface.
Although the runtime is longer because of the involved I/O accesses,
this can be useful for the post-validation of the algorithm instance showing the best performance,
without having to change the code.%

\paragraph{Interface with \irace{} for Automated Algorithm Configuration}

With the approach described previously, the performance of an algorithm instance can be computed
on a given benchmark with the use of a single binary,
without much computation time or memory overhead.
Thanks to the utility features provided by \paradiseo{}, it is straightforward
to expose the meta-algorithm and the problem interfaces
as parameters to the executable (either as parameter files or as command line arguments).
This allows for an easy binding with most automated algorithm configuration tools.

Several automated algorithm configuration tools have been developed in the last decade,
among which one of the most used is \irace{}~\cite{LopDubPerStuBir2016irace}.
\paradiseo{} provides a way to easily expose a foundry interface as an \irace{} configuration file,
as shown in Listing~\ref{lst:irace}.

\begin{lstlisting}[language=C++,label=lst:irace,caption={Excerpt of code for exposing a \paradiseo{} interface to \irace{}.}]
// Using Paradiseo parameters:
eoParser parser(argc, argv, "interface for irace");
auto crossover_p = parser.getORcreateParam<size_t>(
  /*default=*/0, "crossover",
  /*help=*/"The crossover operator", /*flag=*/'c',
  /*help section=*/"Operator Choice", /*required=*/true);
// [...] assemble a foundry [...]
// Print the irace's configuration file for this binary:
std::cout << "# name\t switch\t type\t range\n";
// We only need the parameter(s) and the foundry itself:
print_irace(mutation_rate_p, foundry.mutation_rates,
  std::cout);
print_irace(    crossover_p, foundry.crossovers    ,
  std::cout);
// Any other operator within the foundry [...]
/* This will output something like:
# name       switch             type range
mutationrate "--mutation-rate=" i    (0,4)
crossover    "--crossover="     c    (0,1,2,3,4,5,6,7)
[etc.] */
\end{lstlisting}
With this setting, it has been possible to conduct a large scale algorithm design study%
~\cite{AzizAlaoui2021}
,
involving \irace{} configuring a {\em FastGA} algorithm family (cf. Figure~\ref{fig:FastGA})
solving a W-model problem, using a budget of approximately {\em 1 billion} function evaluations,
in approximately {\em 3 hours} on a single core of a laptop (same setup than Sec.~\ref{sec:fast}).



\section{Conclusions}

This article provides a high-level overview of the \paradiseo{} framework,
a C++ free software which targets the development of modular metaheuristics.
The main feature of \paradiseo{} is its ability to help practitioners to focus on creating solvers while thinking at a higher level of abstraction, thanks to:

    {\bf Utility features:} \paradiseo{} provides a large set of engineering features,
    which very often lack in proof-of-concept frameworks:
    several fine-grained parallelization options, convenient interface features
    (command line argument parsing, state management, useful logs, etc.).
    Having robust implementation of such features is often overlooked by users focusing on the algorithmic part.
    
    {\bf Component-based architecture:} The concept of {\em operator} is at the core of the design of \paradiseo{}.
    It allows for the composition of algorithms, without the overhead of a dynamically loading plugins
    or the rigidity of a monolithic structure.
    Solvers being assembled as a selection of components are also lightweight,
    as it is not necessary to build and carry all the framework's code within the executable binaries.
    
    {\bf Modular algorithm models:} \paradiseo{} provides several mod\-ules targeting different algorithm paradigms
    ---probably one of the largest footprints among active frameworks.
    Practitioners can easily design new algorithms which differ in some operators,
    hybridize algorithms, or even add new algorithm templates using existing operators.
    
    {\bf Algorithm design:} \paradiseo{} focuses on providing a very large design space to the practitioner.
    Thanks to its fast computations, large-scale design experiments can be addressed.
    Combined with its features dedicated to automation,
    \paradiseo{} algorithm designers who want to test a new operator can easily focus on small code changes
    and rapidly check their efficiency and how they interact with other operators in a given paradigm.
    Problem solvers can sort out algorithm instances that work best.

As future works, 
we plan to improve the algorithm design automation features,
merge more state-of-the-art modular designs,
and enhance the overall user experience.

One of the main impediments to a more widespread use of the framework is that the learning curve for getting started is too steep.
Solving this problem is the main objective of \logo{PyParadiseo}, a module currently under development, that will expose interfaces to \paradiseo{} in Python, 
to facilitate the interoperability with external solvers, statistics program or machine learning frameworks.

\section*{Acknowledgments}
During 22 years, \paradiseo{} has been developed by more than 50 people, with the support of the following institutions:
Inria,
University of Lille,
University of the Littoral Opal Coast,
Thales,
École Polytechnique,
University of Granada,
Vrije Universiteit Amsterdam,
Leiden University,
French National Centre for Scientific Research (CNRS),
French National Agency for Research (ANR),
Fritz Haber Institute of the Max Planck Society,
Center for Free-Electron Laser Science,
University of Angers,
French National Institute of Applied Sciences,
Free University of Brussels, Pasteur Institute.

\bibliographystyle{plain}
\bibliography{ARXIV}

%
%

\end{document}